\documentclass[11pt,notitlepage,letterpaper]{article}
\usepackage[utf8]{inputenc}

\usepackage{amsfonts}
\usepackage{eurosym}
\usepackage{lscape}
\usepackage{amssymb}
\usepackage{amsmath}
\usepackage{amsthm}
\usepackage{booktabs, dcolumn} 
\usepackage{graphicx} 
\usepackage{float}
\usepackage{array} 
\usepackage{caption} 
\usepackage{latexsym} 
\usepackage{color}
\usepackage{cite} 
\usepackage{bbold}
\usepackage{array} 
\usepackage{enumerate} 
\usepackage{appendix}
\usepackage{subfigure} 
\usepackage[round]{natbib} 
\usepackage{setspace}
\usepackage{comment}
\usepackage{parskip}
\usepackage{tikz-qtree}
\usepackage{url}
\usepackage{adjustbox}
\usepackage{authblk} 
\usepackage{hyperref}

\theoremstyle{definition}

\theoremstyle{remark}
 
\makeatletter 
\def\@maketitle{ \newpage \null \vskip 2em \begin{center} \let
	\footnote \thanks {\LARGE \@title \par} \vskip 0.5em {\normalsize \lineskip
		.5em \begin{tabular}[t]{c} \large \@author  \end{tabular}\par} \vskip 0.2em
	{\normalsize \@date} \vskip 0.5em {\normalsize \href{}{}}
	\end{center} \par \vskip 1.5em} 
	\makeatother

\usepackage{chngcntr}
\counterwithin{table}{section}
\counterwithin{figure}{section}
\usepackage{rotating}

\setcitestyle{round,aysep={},yysep={;}}
\begin{document}

\author[1]{Sidharth Rony\thanks{Email:
	sidharthrony@gmail.com}}

\author[1]{Jack Patman}

\affil[1]{Royal Holloway, University of London.} 

\title{Standard Occupation Classifier - A Natural Language Processing Approach \thanks{We
	are very grateful to comments from faculty and colleagues in the Royal Holloway.}} 
 \date{February 15, 2024}
\maketitle

\maketitle

\begin{abstract}
  Standard Occupational Classifiers (SOC) are systems used to categorize and classify different types of jobs and occupations based on their similarities in terms of job duties, skills, and qualifications. Integrating these facets with Big Data from job advertisement offers the prospect to investigate labour demand that is specific to various occupations. This project investigates the use of recent developments in natural language processing to construct a classifier capable of assigning an occupation code to a given job advertisement. We develop various classifiers  for both UK ONS SOC and US O*NET SOC, using different Language Models. We find that an ensemble model, which combines Google BERT and a Neural Network classifier while considering job title, description, and skills, achieved the highest prediction accuracy. Specifically, the ensemble model exhibited a classification accuracy of up to 61\% for the lower (or fourth) tier of SOC, and 72\% for the third tier of SOC. This model could provide up to date, accurate information on the evolution of the labour market using job advertisements.
	
 \vspace{2cm}
	\vspace{0.5cm}


\end{abstract}
\newpage

\section{Introduction}

The rise of online job boards has created a vast and valuable data source for economists studying labour market trends. These platforms provide insights into job postings, helping researchers analyse demand for specific occupations and skill sets. Such data is essential for understanding workforce trends, wage determination, and the effects of technological change on employment \citep{vassilev2021s}.  

However, using this data for large-scale analysis requires a standardized system to classify job postings accurately. To extract meaningful insights, job advertisements—comprising job titles, descriptions, and skill requirements—must be mapped to established occupational categories. One of the primary challenges in leveraging online job postings for large-scale labour market analysis is ensuring data quality. Job advertisements often contain inconsistencies that can hinder automated classification processes. These inconsistencies may include typographical errors, ambiguous or vague descriptions, and the frequent use of informal or non-standardized language. Such variations in textual data complicate the extraction of meaningful insights, as machine learning models may struggle to interpret or categorize job postings accurately \citep{vassilev2021s}. The presence of noise in job advertisements necessitates advanced natural language processing techniques capable of handling unstructured data effectively.

In addition to data quality concerns, another significant challenge arises from the dynamic nature of occupational classification systems. Frameworks such as the Standard Occupational Classification (SOC) codes undergo periodic revisions to reflect evolving labour market trends, technological advancements, and emerging job roles. These updates introduce complexities for automated classification models, which must be adaptable to integrate new occupational categories seamlessly while maintaining accuracy in classification. A rigid classification system that fails to account for these periodic modifications risks becoming obsolete, potentially leading to misclassification and diminished analytical reliability. Therefore, any automated solution must incorporate mechanisms for continuous learning and adjustment to ensure alignment with the latest classification standards \cite{lima_bakhshi_2018}.  

Recent advancements in machine learning (ML), particularly in natural language processing (NLP), offer promising solutions to these challenges. NLP techniques, such as word embeddings, help computers interpret job postings by numerically representing words and their meanings. Our research builds on these advancements by applying Bidirectional Encoder Representations from Transformers (BERT) to automate job classification.  

Our approach recognizes that job advertisements consist of three key components: job title and job description, which contains skills required for a job —each offering distinct insights. To improve classification accuracy, we developed three specialized ML models: 

\begin{itemize}
    \item A job title model that captures broad occupational categories. 
    \item A job description model that analyses detailed job duties and qualifications. 
    \item A skills model that identifies and classifies required competencies.  
\end{itemize}

By training separate models for each component, we enhance the accuracy of classification, ensuring that each model learns the most relevant linguistic patterns. To further improve performance, we use an ensemble learning technique, which combines the predictions from all three models. This approach leverages the unique strengths of each model—for example, the job title model may excel at broad categorization, while the skills model better identifies specific qualifications. By integrating their outputs, we achieve more precise and reliable job classification \citep{dietterich2000ensemble}.

This study introduces a novel approach that incorporates skills alongside job titles and descriptions within the classification process. By leveraging a neural network to classify skills extracted from job descriptions, our model achieves superior accuracy across all classification tasks. This integration of skill-based information represents a significant contribution to the field and has the potential to revolutionize the analysis of online job advertisements. The following sections of the paper will delve deeper into the specifics of our methodology, showcasing the improved accuracy achieved through our approach.

\section{Background}
This section aims to provide a comprehensive understanding on language processing models and the challenges that this approach implies.

\subsection{Standard Occupational Classification }

Standard Occupational Classification (SOC) codes, such as those employed in the United States (O*NET) and the United Kingdom (ONS SOC), represent meticulously constructed taxonomies for categorizing jobs.  These classification systems undergo rigorous manual reviews at periodic intervals to guarantee their accuracy and continued relevance within the evolving labour market.
A defining characteristic of SOC codes is their hierarchical structure.  The initial digit or group of digits denotes a broad occupational category, while subsequent digits progressively refine the classification into more specific occupational subsets.  The complete code, typically consisting of four digits, signifies the most granular level of detail within the occupational taxonomy.

\begin{figure}[H]
\begin{tikzpicture}
\Tree [.{5 - Skilled Trades Occupations} [.{52 - Skilled Metal Trade} 
[.{523 - Vehicle Trade } [.{5235 - Aircraft Maintenance} ]] [.{524 - Electrical Trade} {...} ] ] 
[.{53 - Skilled Construction } {...} ] ]
\end{tikzpicture}
\caption{ UK ONS SOC Classifier structure for an aircraft maintenance occupation.}
\end{figure}
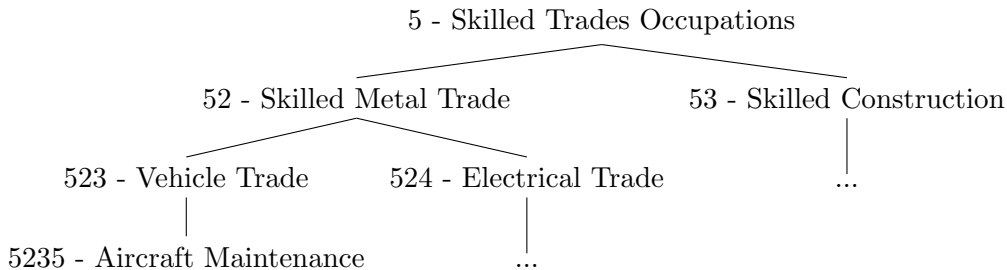

For the purpose of this project, we will be concerning ourselves with two standards - the UK Office for National Statistics SOC SOC, and the US Department of Labour O*NET SOC. These are structured respectively as follows:

\begin{figure}[H]
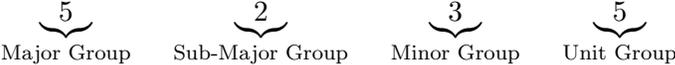

\[
\underbrace{5}_{\text{Major Group}} \hspace{0.5cm}
\underbrace{2}_{\text{Sub-Major Group}} \hspace{0.5cm}
\underbrace{3}_{\text{Minor Group}} \hspace{0.5cm}
\underbrace{5}_{\text{Unit Group}} \hspace{0.5cm}
\]
\caption{Subset of the ONS SOC 2020 classification demonstrating the taxonomy.}
\end{figure} 

\begin{figure}[H]
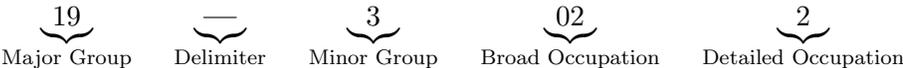

\[
\underbrace{19}_{\text{Major Group}} \hspace{0.5cm} 
\underbrace{\textbf{---}}_{\text{Delimiter}} \hspace{0.5cm} 
\underbrace{3}_{\text{Minor Group}} \hspace{0.5cm}
\underbrace{02}_{\text{Broad Occupation}} \hspace{0.5cm}
\underbrace{2}_{\text{Detailed Occupation}} \hspace{0.5cm}
\]
\caption{US O*NET SOC Classifier structure for a survey research occupation.}
\end{figure}

Several factors contribute to the complexity of mapping available employment data back to corresponding SOC codes.  Firstly, the hierarchical nature of the classification system necessitates assigning data points to the most appropriate level within the hierarchy.  This can be challenging, particularly for job descriptions that encompass a broader range of activities or bridge the boundaries between different occupational categories.

Secondly, the sheer number of distinct classifications poses a significant hurdle.  For instance, O*NET 2019 encompasses 1016 unique classifications, while ONS SOC 2020 features 412 distinct categories.  Effectively mapping data points to such a vast number of classes requires robust classification algorithms capable of handling this level of complexity.

Thirdly, the periodic review and revision process inherent to SOC codes introduces an additional layer of difficulty.  As these classification systems are manually updated every ten years to reflect changes within the labor market, data collected under previous versions of the code (e.g., ONS SOC 2010) may not directly translate to the newly revised categories (e.g., ONS SOC 2020).  This necessitates the development of strategies to bridge these temporal gaps and ensure accurate mapping of historical data to the most relevant classifications within the current coding system.

To effectively address these challenges, our research explores the application of advanced classification algorithms, such as those leveraging deep learning techniques, to enhance the accuracy and efficiency of mapping employment data to SOC codes.  Furthermore, we investigate methods for handling the hierarchical structure of the classification system and incorporating techniques to account for the revisions introduced during periodic reviews.  By overcoming these hurdles, we aim to facilitate more precise and nuanced analyses of labor market trends and dynamics.

\begin{figure}[H]
\centering
	\includegraphics[width=9cm]{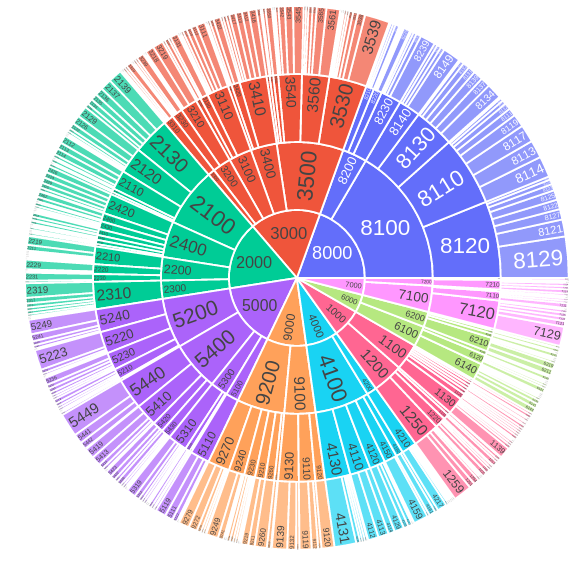}
	\caption{The SOC2020 taxonomy - demonstrating the significant stratification of the final label classifications. Different colours are representative of different 'major group' occupational codes.}
	\label{fig:soc2020tax}
\end{figure}

Figure \ref{fig:soc2020tax} depicts the hierarchical structure of the Standard Occupational Classification (SOC) 2020 system. The use of color differentiates between the various "major group" occupational codes, representing the broadest classifications within the taxonomy.  These major groups are further subdivided into more granular categories through subsequent levels of the hierarchy.

The figure serves to visually illustrate the significant stratification present within the final label classifications of the SOC 2020 taxonomy. This stratification highlights the challenge of accurately classifying job advertisements at the most detailed level (typically the four-digit code) due to the vast number of distinct occupational categories encompassed by the system.

Economists studying labour market trends and dynamics can leverage this figure to gain a better understanding of the granularity inherent in occupational classification systems like SOC 2020.  This knowledge is crucial for interpreting and analysing employment data categorized using these codes. 

\subsection{Literature Review}

Assigning Standard Occupational Classification (SOC) codes to job advertisements is crucial for labour market analysis and job matching. This review examines existing methods for automating this process, focusing on the provided references.

Several approaches leverage textual analysis of job postings. \cite{atalay2020evolution} utilize a Continuous Bag-of-Words (CBOW) model to identify job titles with similar meanings in newspaper ads, achieving an  of  53\% for 4-digit O*NET SOC codes using pre-defined title-to-SOC code mappings. \cite{turrell2019transforming}  utilizes a rule-based approach to assign 3-digit SOC codes. It leverages job titles, sector information, and job description text, acknowledging the trade-off between achieving finer classifications (more digits in the SOC code) and maintaining accuracy. While the exact accuracy for all 3-digit SOC codes isn't explicitly stated, the paper offers valuable insights. By comparing its performance to the Office for National Statistics' (ONS) labeling system, the study reveals that the ONS system struggles with a significant portion of job postings due to unclear titles. The vacancy postings method, however, significantly improves accuracy by incorporating descriptions. When tested on vacancies the ONS system could confidently label, the vacancy postings method achieved a 91\% match in assigning the same 3-digit SOC code. This suggests the method offers a more robust and accurate solution compared to relying solely on job titles.

Supervised machine learning also shows promise. \cite{gweon2017three}, though not using job ads directly, demonstrate that a modified nearest neighbor approach can achieve high accuracy for occupation coding based on survey data from Germany, suggesting its potential application to job ads. \cite{boselli2018wolmis} present the WoLMIS system, which utilizes supervised machine  methods to classify web job vacancies according to ISCO (International Standard Classification of Occupations) codes, a close equivalent to SOC. Their evaluation highlights the effectiveness of n-gram features extracted from both job titles and descriptions, with SVMs achieving a best f1-score of 0.91 for classifying vacancies into the first level (9 distinct occupation groups) of the ISCO taxonomy.

In this paper we leverage the advance capabilities of BERT transformer to classify raw job advertisements to their respective UK SOC 2010, UK SOC 2020 and O*NET SOC 2019.

\subsection{Language Models}

Online job advertisements offer a wealth of information about labor market trends. However, unlike structured data sets commonly used in economic analysis, job postings are inherently unfiltered. This free-text nature presents a challenge: how to extract reliable information about occupations from potentially noisy data.

Two main approaches exist to address this challenge:

\begin{itemize}
    \item Data Preprocessing: This involves meticulously cleaning the data by removing irrelevant information, correcting typos, and standardizing language. While effective, this approach can be time-consuming and resource-intensive.

\item  Noise-Resistant Models:  This approach utilizes machine learning models specifically designed to handle noisy data. These models can learn from the raw, unfiltered job advertisements, mitigating the need for extensive preprocessing.
\end{itemize}

Early attempts at representing text numerically (embedding) relied on methods like tf-idf, word2vec, and GloVe. While these techniques offer some level of success, they  often struggle to capture the nuances of language, particularly in the context of job advertisements.

These traditional methods fall short in two cases. First when the context is limited.  These models primarily focus on the individual words themselves, neglecting the broader context within a sentence or job description. This can lead to misinterpretations, especially for words with multiple meanings (polysemous terms). For instance, the word "skill" could refer to technical skills, interpersonal skills, or even a particular fish depending on the context. Secondly when there is a large corpora involved. Job advertisements often contain a vast amount of text (corpus). Traditional embedding methods struggle to handle this complexity, especially when dealing with polysemous terms that require disambiguation based on the surrounding text.

Recent advancements in machine learning, particularly the development of transformer-based Large Language Models (LLM) like BERT, XLNet, and GPT-3, have revolutionized the field of large document analysis. LLMs are a class of powerful machine learning models trained on massive amounts of text data. This training allows them to understand complex relationships between words and sentences. These models offer significant advantages over traditional techniques: Transformer-based models excel at working with raw, unprocessed text. This eliminates the need for extensive data cleaning, saving time and resources. These models are designed to understand the relationships between words within a sentence and the broader context of a document. This allows them to accurately interpret polysemous terms and capture the true meaning within a job advertisement.

Given the remarkable success of BERT and its powerful variants, our research focuses on this family of embedding models. BERT, in particular, has established itself as a leading force in text classification tasks, including occupational classification from job advertisements. We examine the performance of BERT and its variants  (\cite{devlin2018bert}, \cite{yang2019xlnet}, \cite{brown2020language}) in detail.  Our goal is to leverage the strengths of BERT-based models to handle the noisy nature of job advertisement data and achieve superior performance in occupational classification tasks.

\subsection{Large Document Analysis}

One of the key limitations associated with BERT and its variations is the restriction on input sequence length, typically capped at 512 tokens. This necessitates strategies for handling job advertisements exceeding this limit.

The prevailing approach involves truncating a portion of the input text that allocates a quarter of the available tokens (e.g., 128 tokens for a 512-token sequence) to the initial portion of the advertisement and the remaining three-quarters to the concluding section \citep{SunChi2019HtFB}. This prioritizes the introductory information, often containing concise descriptions aimed at capturing the target audience's attention, and the concluding remarks that typically detail key requirements of the advertised position. The effectiveness of various truncation strategies, particularly considering the potential influence of domain context on optimal truncation locations, will be investigated and reported upon in our research.

For a small minority of job advertisements exceeding 512 tokens, a hierarchical approach can be considered. This approach involves two potential methods, both of which segment the lengthy document into n sub-documents, each with a maximum of 512 tokens. Each sub-document is then encoded separately using the chosen embedding model (e.g., BERT).

In the first hierarchical approach, max pooling is applied to the embeddings derived from the sub-documents. Max pooling identifies and retains the most significant sub-tokens (those with the highest attention scores) to generate a condensed output that meets the 512-token limit.

Alternatively, the second hierarchical approach involves performing separate predictions on each sub-document. The individual model outputs are then combined through a form of ensemble learning to arrive at a final prediction for the entire document.

While a hierarchical approach offers a solution for exceptionally long documents, we opt to prioritize exploring truncation strategies for several reasons. Firstly, the number of job advertisements exceeding 512 tokens is relatively small. Secondly, the computational burden associated with managing very long documents is significant due to the quadratic complexity inherent in fine-tuning the self-attention layers within BERT.  Therefore, focusing on efficient truncation strategies presents a more practical solution for the majority of job advertisement text data used in occupational classification tasks.

This analysis highlights the trade-offs between various approaches for hierarchical classification of occupations with BERT. We will leverage LCPN due to its balance between accuracy and efficiency, while acknowledging the limitations of current objective functions in exploiting the full potential of hierarchical taxonomies within deep learning models. Future research can explore these limitations and delve deeper into alternative evaluation metrics specifically designed for hierarchical classification tasks.

\subsection{General Overview}

I first collected the job postings from the UK between March 2020 and July 2020 by web scrapping from Indeed\footnote{\url{https://uk.indeed.com/?r=us}}. A representative sample was carefully labelled with the corresponding codes from three prominent classification systems: UK SOC 2010, UK SOC 2020, and O*NET SOC 2019. During the training phase (the process of feeding data to the models to enable them to learn), various NLP models, including BERT, DistilBERT, RoBERTa, and DeBERTa, were employed to analyse both job titles and descriptions. Additionally, a neural network classifier was utilized to handle the classification of skills. All the three title, description and skills from description were used to train the model that can classify job advertisements. Finally, an ensemble model, which combines the predictions from the individual models, was constructed to enhance classification accuracy.

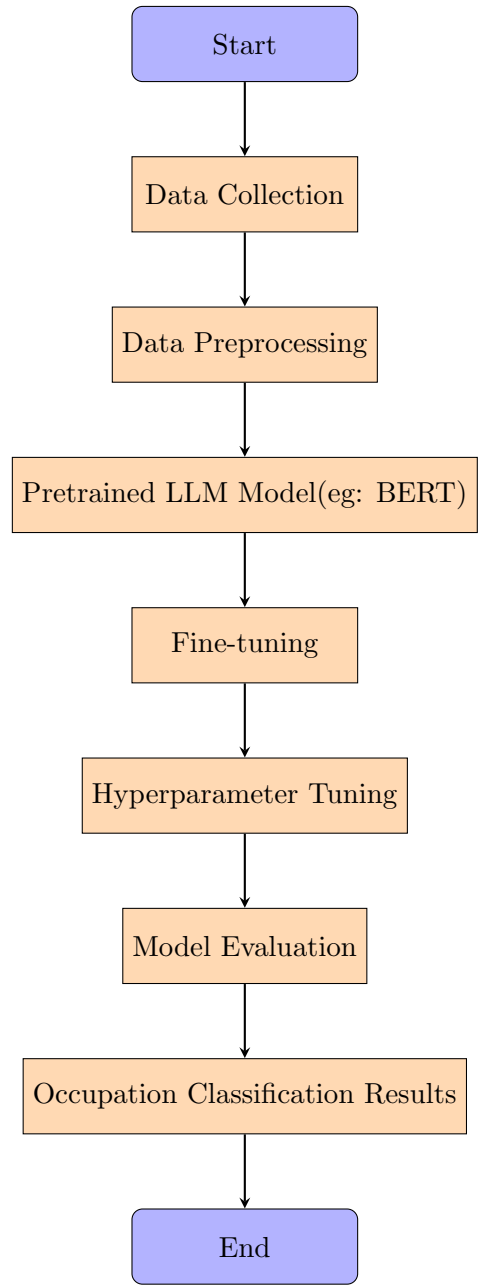
\begin{figure}[H]
\tikzstyle{startstop} = [rectangle, rounded corners, minimum width=3cm, minimum height=1cm,text centered, draw=black, fill=blue!30]
\tikzstyle{process} = [rectangle, minimum width=3cm, minimum height=1cm, text centered, draw=black, fill=orange!30]
\tikzstyle{arrow} = [thick,->,>=stealth]
\centering
\begin{tikzpicture}[node distance=2cm]

\node (start) [startstop] {Start};
\node (data) [process, below of=start] {Data Collection};
\node (preprocess) [process, below of=data] {Data Preprocessing};
\node (bert) [process, below of=preprocess] {Pretrained LLM Model(eg: BERT)};
\node (finetune) [process, below of=bert] {Fine-tuning};
\node (hyperparam) [process, below of=finetune] {Hyperparameter Tuning};
\node (evaluate) [process, below of=hyperparam] {Model Evaluation};
\node (results) [process, below of=evaluate] {Occupation Classification Results};
\node (end) [startstop, below of=results] {End};

\draw [arrow] (start) -- (data);
\draw [arrow] (data) -- (preprocess);
\draw [arrow] (preprocess) -- (bert);
\draw [arrow] (bert) -- (finetune);
\draw [arrow] (finetune) -- (hyperparam);
\draw [arrow] (hyperparam) -- (evaluate);
\draw [arrow] (evaluate) -- (results);
\draw [arrow] (results) -- (end);

\end{tikzpicture}
\caption{Workflow of Occupation Classification}
\end{figure}
In our case,  each LLM is "fine-tuned" for the specific task of occupation classification. This fine-tuning involves feeding the LLM a labeled dataset of job advertisements, where each advertisement is associated with its corresponding occupation code. The dataset is typically divided into a training set used to train the model and a test set used to evaluate its performance. During fine-tuning, we optimize the learning process by carefully selecting critical hyperparameters such as the number of epochs (training iterations), learning rate (speed of learning), and batch size (number of training examples processed at once). These hyperparameters are crucial for achieving optimal performance. We evaluate the model's effectiveness using metrics like accuracy or cost functions. By successfully fine-tuning the LLM, we essentially equip it with the ability to classify skills within job advertisements, which in turn, allows for accurate classification of the advertised occupation itself. This approach offers a robust and efficient solution for automating occupation classification from online job advertisements.

\section{Data}\label{datasetorig}
\label{datasetorig}

This section details the data acquisition and pre-processing steps undertaken to prepare a comprehensive dataset for training and evaluating our occupational classification model with BERT.

\subsection{Data Collection}
\label{dataorig}

The cornerstone of our dataset is a collection of occupation standards maintained by national statistical agencies (e.g., Office for National Statistics (ONS) in the UK , Bureau of Labor Statistics (BLS) in the US). These standards provide official job titles and their corresponding occupational classification codes, serving as a foundation for our labeling process.

In addition to these official sources, we leverage publicly available datasets that significantly expand the range of job title variations considered. The  "SOC Coding Compendium" offers a comprehensive set of roughly 30,000 job title variations mapped to minor occupation groups within the ONS SOC 2020 and SOC 2010 classification systems.  Similarly, the "O*NET Title Set" provides access to an additional 52,907 alternate job titles specifically associated with the O*NET 2019 SOC taxonomy.

While these supplementary datasets enrich our model training process by exposing the model to a broader spectrum of job title phrasings, our primary focus is on developing a classifier adept at handling the inherent noise and idiosyncrasies present in real-world job advertisements.

A critical component of this research involves the creation of a comprehensive dataset for training and evaluating our occupational classification model. To achieve this, we opted to leverage a rich source of real-world job postings: the online job board Indeed.com.  Through a web scraping process, we were able to gather a substantial corpus encompassing approximately 500,000 job advertisements posted within the UK over a six-month period in 2020.

This decision to utilize Indeed.com as the primary data source is grounded in several key considerations.  Firstly, Indeed.com is a prominent online job board attracting a vast array of employers across various industries and occupational categories.  This ensures that the collected data reflects a broad spectrum of job postings, encompassing a diverse range of occupations and sectors within the UK labor market.

Secondly, the sheer volume of data obtained (500,000 job advertisements) allows for the development of a robust and statistically significant training dataset.  A larger dataset provides the model with a wider range of examples to learn from, enhancing its ability to generalize and perform accurately on unseen data.

Thirdly, by focusing on job advertisements posted within a specific time frame (six months in 2020), we mitigate the potential influence of temporal variations in job postings.  Since job descriptions and required skills may evolve over time, restricting the data collection window helps maintain consistency within the dataset and reduces the risk of introducing biases associated with temporal trends in job advertisements.

The resulting corpus of job advertisements serves as the foundation for our research.  

\subsection{Data Preparation}

We here  address the challenge of transforming this vast collection of job advertisement text data into a format suitable for training a machine learning model.  This critical step involves creating a labelled dataset where each job advertisement is assigned a corresponding occupational code.

Given the substantial size of the raw data (500,000 job advertisements), directly training a model on this entire corpus would be computationally expensive and potentially inefficient.  To address this, we employ a random sampling technique to extract a representative subset of 8,511 job advertisements from the original data. 

The next critical step involves meticulous hand-labeling of each job advertisement within the chosen subset.  This labeling process is entrusted to research assistants(RA) with detailed instructions .  Each RA assigned a corresponding occupational code from all three classification systems (ONS SOC 2020, ONS SOC 2010, and O*NET 2019) to every job advertisement in the subset.  This multi-code labeling approach allows us to not only train the model for the most recent classification system (ONS SOC 2020) but also facilitates the application of the model to analyze historical data coded under previous versions of the system (ONS SOC 2010).

The meticulous nature of this hand-labeling process is paramount.  By ensuring the accuracy and consistency of the assigned occupational codes, we establish a reliable reference point for training the model.  This labeled dataset serves as the cornerstone upon which the model learns to identify the key characteristics and linguistic patterns within job advertisement text data that are indicative of specific occupations.

Real-world job advertisements are inherently noisy and exhibit a high degree of variation.  Job titles may not precisely align with standardized occupational classifications, and job descriptions may employ informal language or industry-specific jargon.  The hand-labeling process account for these inherent variations.  By carefully considering the content of each job advertisement and selecting the most appropriate code(s) from the classification systems, the RAs bridge the gap between the sometimes-ambiguous language of job postings and the standardized categories within occupational taxonomies.

Furthermore, the substantial size and representative nature of the labelled dataset (8,511 job advertisements) allows the model to learn from a wide range of examples encompassing these variations.  This exposure to the complexities of real-world job advertisement text data enhances the model's ability to handle noise and generalize effectively to unseen data, ultimately leading to more robust and accurate occupational classifications.

\subsection{Data Summary}

\begin{figure}[H]\label{titlefig}
	\includegraphics[width=13cm]{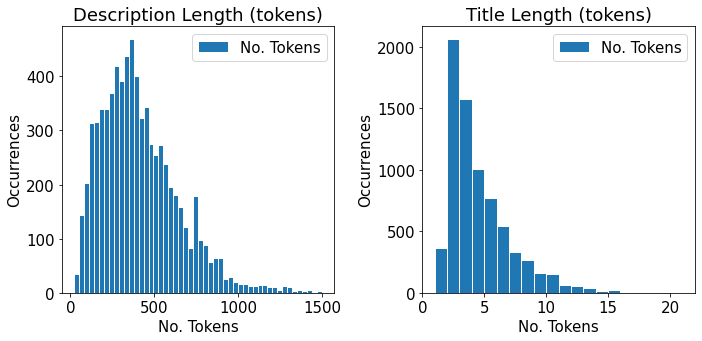}
	\caption{Distribution of Job Advertisement and Title Token Length.}
	\label{fig:largeDocAnalysis}
\end{figure}

A key consideration is the maximum sequence length that BERT can effectively process.  While the standard BERT model is capable of handling sequences up to 512 tokens, exceeding this limit necessitates preprocessing steps to ensure optimal model performance. Our analysis of the collected job advertisement data reveals that a small portion (28.8\%) surpasses the 512-token limit imposed by BERT.  Figure \ref{fig:largeDocAnalysis}  visually depicts the distribution of job advertisement and title token lengths within our dataset.  This finding underscores the importance of incorporating preprocessing techniques to address job advertisements exceeding the BERT input limit.

Several strategies can be employed to circumvent the 512-token constraint.  One approach involves truncating overly long job advertisements to a predefined maximum length (e.g., 512 tokens).  While this method is straightforward, it may lead to the loss of potentially valuable information contained within the truncated text.  Alternatively, a more nuanced approach involves segmenting the job advertisement text into smaller chunks that fall within the acceptable length limit.  These segments can then be fed into the BERT model sequentially, potentially with additional context information incorporated to bridge the gaps between segments.  Another strategy leverages techniques such as sentence transformers, which encode entire sentences into fixed-length vectors, enabling the model to capture the essence of the job advertisement without exceeding the token limit.

The selection of the most suitable preprocessing approach hinges on a careful evaluation of the trade-offs between information retention and computational efficiency.  Truncation offers simplicity but risks discarding valuable details.  Segmentation necessitates more complex model architectures to handle segmented inputs, while sentence transformers may require additional training data specifically tailored for this task.

By effectively addressing the length constraints inherent in job advertisement text data, we can ensure compatibility with the BERT model and optimize its ability to extract relevant information for accurate occupational classification.

\section{Methodology}

We developed and trained three distinct models aimed at classifying job titles, job descriptions, and skills based on their respective labelled SOC codes. Our approach involved creating separate models for each of these components to effectively capture the nuances and unique characteristics of each aspect of the job advertisement. The ensemble method was employed to enhance the overall performance and prediction accuracy of our classification system. By combining the outputs of the three individual models, we sought to leverage their complementary strengths, thus achieving more robust and reliable results.

While job titles and descriptions were readily available within the job advertisements, the skills associated with each job were extracted using a specifically designed algorithm \citep{karlisskill}. This algorithm enabled us to identify and capture the relevant skills present in the job postings, which contributed significantly to the comprehensive understanding and categorization of each job's requirements.

By adopting this multi-model approach and incorporating a skill extraction algorithm, we aimed to create a powerful and versatile system capable of accurately assigning SOC codes to job titles, descriptions, and extracted skills. The combination of these elements allowed us to gain valuable insights into the diverse attributes of job advertisements and facilitate more informed decision-making in various occupational domains.

\begin{figure}[H]
\centering
	\includegraphics[width=16cm]{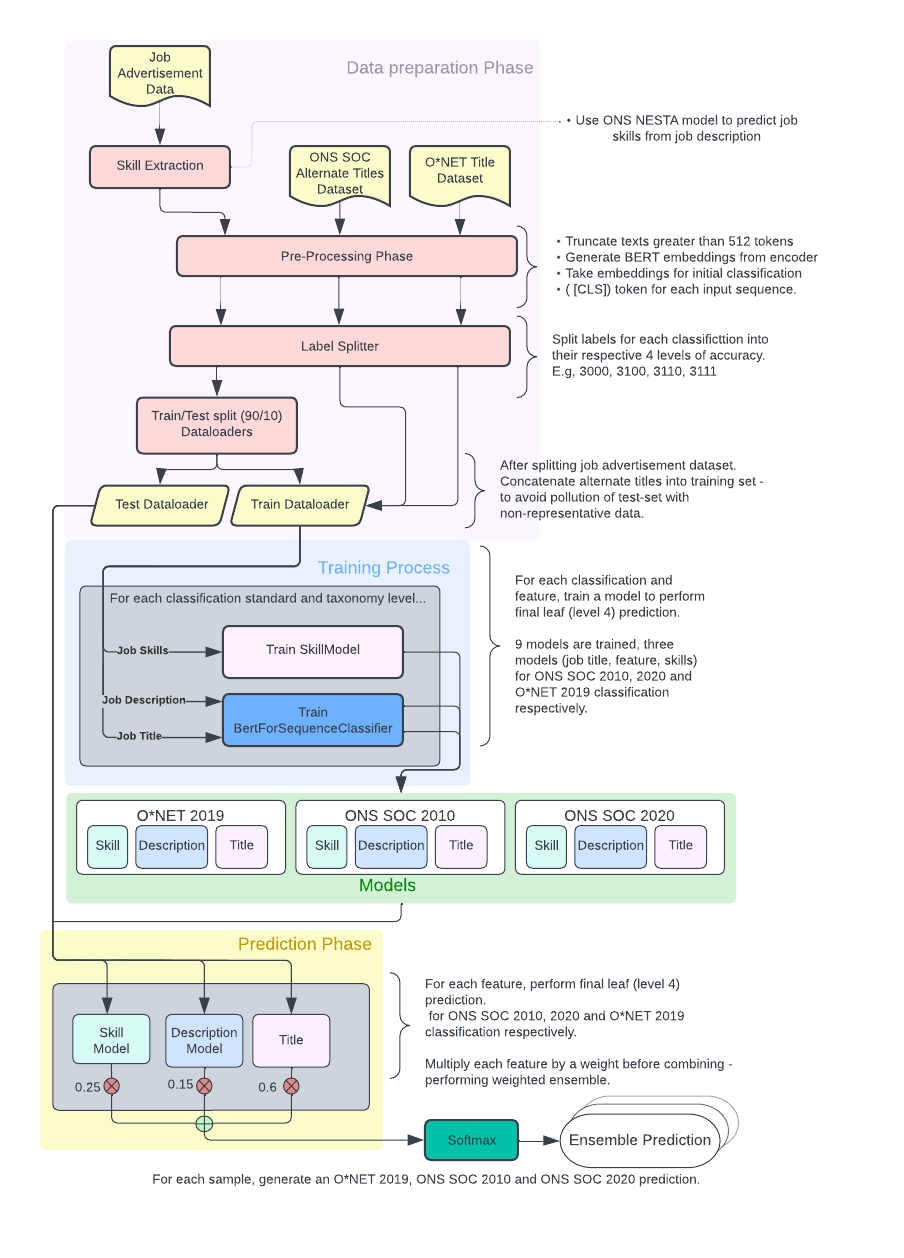}
	\caption{Brief overview of the model training and prediction process used for the final results table}
	\label{fig:finalmodelfig}
\end{figure}

Figure \ref{fig:finalmodelfig} depicts the workflow of the final model used to generate the results presented in subsequent sections.  The figure visually illustrates the separate training processes for the job title, job description, and skills extraction models.  It then showcases how the outputs from these individual models are combined within the ensemble learning framework to arrive at the final predicted occupational code (SOC code) for a given job advertisement.

\subsection{Preprocessing}\label{preprocessing}
A key benefit of utilizing BERT, compared to traditional embedding models, lies in its capability to achieve accurate classifications even when presented with text containing stop words, punctuation, and other elements often categorized as noise ({\cite{AlzahraniEsam2021HDTT}}, {\cite{EkAdam2020HdPA}}). This advantage aligns intuitively with the inherent structure of the self-attention mechanism employed by modern transformer models.  The self-attention principle ensures that only words significantly impacting the final classification receive weight (attention) during the processing.

While we will conduct experiments involving the removal of less frequent noise elements during model development to further validate this observation, standard Natural Language Processing (NLP) pre-processing steps like stop-word removal will not be routinely applied. Our preprocessing approach focuses on addressing specific elements within the job advertisement text data that might hinder the model's learning process.  These techniques aim to improve the model's ability to focus on the most relevant information for accurate occupational classification.  A summary of the implemented preprocessing steps is provided in the table below (Table \ref{tab:preprocessing}).

\begin{table}[H]
	\centering
 \caption{\label{tab:preprocessing} Preprocessing Techniques}
	\begin{tabular}{ |p{4cm}|p{8cm}|  }
		\hline
		\multicolumn{2}{|c|}{Preprocessing} \\
		\hline
		\textbf{Approach} & \textbf{Example} \\
		\hline
		Split 'or' phrases  & Cleaner/maid $\to$ cleaner maid \\
		\hline
		Remove text in brackets  & Night care assistant (Kirby House) $\to$ Night care assistant \\
		\hline
		Remove hour ranges & Customer service 10hrs $\to$ Customer service  \\
		\hline
		Remove salary ranges & Carer £9.20- £10.50 per hour $\to$ Carer  \\
		\hline
		Truncate after hyphen & Installation assistant-CSCS card holder' $\to$ Installation assistant  \\
        \hline		
	\end{tabular}
\end{table}

Lemmatization and stemming aim to reduce words to their base forms. Lemmatization considers grammatical context (e.g., "running" to "run"), while stemming uses predefined rules (e.g., "running" to "runn"). The decision to employ lemmatization or stemming techniques hinges on a deeper understanding of the tokenization process utilized by BERT.  BERT leverages a specific tokenization method known as WordPiece tokenization.  During the training process for WordPiece tokenization, individual words are initially broken down into their constituent letters, which then form the starting vocabulary.  Subsequently, bigrams (pairs of consecutive letters) are constructed from this vocabulary, and a score is assigned to each bigram based on its frequency compared to the individual letter frequencies.  This process iteratively expands the vocabulary by creating permutations of letters with corresponding frequency scores.  During tokenization, the WordPiece sub-token vocabulary performs a partial match on the input token, aiming to identify the longest possible partial match.  This match is then converted into a separate token.  The process continues by seeking the second-longest partial match and so on, effectively splitting the original token into sub-tokens of progressively shorter lengths.

Since this matching process occurs at the sub-token level (partial matches), it avoids the potential drawbacks associated with full-token stemming or lemmatization.  By focusing on sub-tokens, WordPiece tokenization can handle variations in word forms without the risk of losing potentially relevant information that might be eliminated through traditional stemming or lemmatization techniques.  Therefore, in the context of our BERT-based approach, lemmatization and stemming are not necessary preprocessing steps.

A previous study by \cite{karlisskill} introduced an NLP model for extracting skill sets from job descriptions. We hypothesize that these extracted skills will be strongly linked to the occupational category of the job advertisement.  Therefore, this project will investigate the effectiveness of incorporating these skills, obtained through \cite{karlisskill}, as an additional feature within our final model for occupational classification. This has the potential to enhance the model's accuracy by leveraging the valuable information embedded within the specific skills demanded for each position.


\subsection{Train, Test Split}

The selection of an appropriate data splitting ratio for training, validation, and testing purposes is crucial for robust model evaluation.  While conventional wisdom often suggests an 80/20 split,  we opted for a 90/10 split in this instance.  This deviation from the typical rule of thumb stems from the  relatively limited availability of our labelled training data.

A larger training dataset allows the model to learn a broader range of features and relationships within the job advertisement text data, ultimately leading to improved generalization and performance on unseen data.  Given the constraints of our labelled dataset, utilizing 90\% of the data for training and validation offers a more robust foundation for model development compared to a more traditional 80/20 split.

Within the chosen 90\% allocation for training and validation, we further partition the data into a held-out test set comprising 10\% of the total data.  This held-out test set serves as a completely unseen dataset for evaluating the final model's generalizability and performance on data it has not encountered during training.

 \begin{table}[H]
 	\centering
 	\begin{tabular}{ |p{3cm}||p{3cm}|p{4cm}| }
 		\hline
 		\multicolumn{3}{|c|}{Initial Model Performances} \\
 		\hline
 		Classification  & Training Set Size & Test Set Size \\
 		\hline
 		ONS SOC 2010  & 7871 & 1929 \\
 		ONS SOC 2020  & 7874 & 1930 \\
 		O*NET 2019  & 7394 & 2127 \\
 		\hline
 	\end{tabular} 
 	\caption{ Training-Test split for each classification. Note minor variance where values including a  '0' label - representing an unknown label for that classification, were filtered out. }
 \end{table}

\subsection{Model}
\label{model}

Our initial analysis used the BERT model to process job advertisement text and create meaningful numerical representations (embeddings) of the data. These embeddings were then fed into different types of neural networks to classify job postings into occupational categories. We
experimented with various neural network structures, adjusting their size, activation functions,
and techniques to prevent overfitting\footnote{Technical details of model parameters are provided in Appendix \ref{model:appendix}}.

One challenge was handling unstable gradients—sudden changes in model learning—which we managed using gradient clipping. To prevent the model from memorizing training data instead of generalizing, we used early stopping, halting training when improvements stopped. We also experimented with dropout, a technique that randomly disables parts of the network during training, and layer normalization, which stabilizes learning by standardizing neuron activations.

For classification, we explored a hierarchical approach called the Local Classifier per Parent Node (LCPN). This method first assigns a broad occupational category and then refines predictions at more specific levels. However, if the first classification was incorrect, errors could propagate through the system. To mitigate this, we introduced confidence thresholds, ensuring that only high-confidence predictions moved to the next stage.

To refine final predictions, we tested three methods for incorporating confidence scores:

\begin{itemize}
    \item Total Average – gives equal weight to all classification levels.
    \item Weighted Average – prioritizes important classifications.

    \item Joint Probability – multiplies confidence scores across levels.

\end{itemize}

Another key improvement involved gradient accumulation, which allows larger batches of data to be processed without requiring excessive computing power (relatively). This method effectively increased the efficiency of training without sacrificing accuracy. We used AdamW, an advanced optimization technique, to adjust model weights efficiently. Additionally, we fine-tuned hyper-parameters—key settings that influence model performance—using Optuna, a specialized tool for automated tuning. Through 100 trials, we optimized parameters like learning rate, dropout rates, and weight decay, ensuring the best balance between accuracy and computational efficiency. To further boost accuracy, we combined multiple models in an ensemble approach, where different classifiers using job titles, descriptions, and extracted skills contributed to the final prediction. This strategy led to a 3-5\% improvement in accuracy—seemingly small, but significant given the scale of job classification.

\section{Results and Analysis}

This section delves into the exploration of various modelling strategies employed to achieve the most accurate and efficient occupational classification system.  We begin by investigating feature engineering techniques, including text truncation for job descriptions and the evaluation of individual features (job titles, descriptions, skills, and official titles) to assess their relative importance in classification accuracy.  Next, we explore pre-processing techniques for job titles and compare the performance of flat and hierarchical classification models.  Furthermore, we leverage the power of pre-trained language models like BERT to generate contextual embeddings from job descriptions, further enriching the feature set for classification.  By systematically evaluating these diverse approaches, we aim to identify the optimal model architecture that balances classification accuracy, computational efficiency, and interpretability – crucial factors for real-world applications in labor market analysis and workforce development initiatives.

In this paper, F1 (Macro) and accuracy are metrics used to evaluate the performance of our classification model. Accuracy simply represents the proportion of correctly classified occupations. F1 (Macro), however, provides a more balanced view, especially when dealing with multiple occupational categories. It calculates an average of the F1 scores across all individual classes, ensuring each category contributes equally to the final score. F1 itself considers both precision (percentage of correctly predicted positives) and recall (percentage of actual positives the model identified) for each class, providing a more comprehensive performance measure.

\subsection{Truncation Strategies Performance}\label{truncstratresult}

In the context of our occupational classification model, a key challenge involved handling lengthy job descriptions while adhering to the model's limited token capacity (512 tokens). To address this constraint, we investigated three truncation strategies: "mixed," "tail," and "head." The "mixed" strategy combined the first 3/4 (384 tokens) and the final 1/4 (128 tokens) of the job description, aiming to capture information from both the beginning and end of the text. Conversely, the "tail" strategy utilized only the last 512 tokens, focusing on the most recent information provided. Finally, the "head" strategy concentrated solely on the initial 512 tokens of the job description.

Table \ref{tab:truncation} summarizes the performance of these strategies on the final level classification task for the ONS SOC 2020 dataset.  As the table reveals, the "head" strategy achieved the most favorable results, with a test F1 (Macro) score of 0.215 and a test accuracy of 49.64\%. This outcome was unexpected, as we initially hypothesized that a combination of the head and tail sections (e.g., "mixed" strategy) might capture the most relevant information for classification.  The observed performance suggests a negative correlation between the inclusion of a larger portion from the tail of the job description and overall model performance. This finding warrants further investigation to understand the specific content within job descriptions that contributes most significantly to accurate occupational classification.

\begin{table}[H]
	\centering
	\begin{tabular}{ |p{2cm}||p{3cm}|p{4cm}| }
		\hline
		\multicolumn{3}{|c|}{Truncation Strategy Performances} \\
		\hline
		Model  & Test F1 (Macro) & Test Accuracy (\%) \\
		\hline
		Mixed  & 0.207 & 48.652 \\
		Tail  & 0.182 & 47.098\\
		Head  & \textbf{0.215} & \textbf{49.637} \\
		\hline
	\end{tabular} 
	\caption{\label{tab:truncation}Truncation model test performances ONS SOC 2020 description - final level classification. Best results emphasized.}
\end{table}

\subsection{Job title Classification} 
In the initial stages of development, we explored the possibility of training a separate predictive model on the "clean" job titles introduced in Section \ref{datasetorig} (refer to original data description). This model was intended for inclusion within the final ensemble strategy.  However, during testing, an interesting phenomenon emerged. While the model trained on clean titles performed comparably to the model trained on noisy job advertisement titles when evaluated on a test set of clean titles, its performance deteriorated significantly when applied to an unclean test set, particularly at more granular levels of the occupational taxonomy.  This observation suggested a potential overfitting issue with the clean title model.  Consequently, we opted for an alternative approach: concatenating the job titles from the original advertisement dataset with the titles derived from the official compendium sets. This strategy aimed to leverage the richness of the original data while mitigating potential overfitting associated with the clean title set.

\begin{table}[H]
	\begin{tabular}{ |p{4cm}|p{3cm}|p{2cm}|p{4cm}|  }
		\hline
		\multicolumn{4}{|c|}{Compendium - Standalone Model Performances}\\
		\hline
		Classification  & Taxonomy Level & Test F1 & Test Accuracy (\%) \\
		\hline
		ONS SOC 2010 & 1 & 0.499 & 53.240 \\
		ONS SOC 2010 & 2 & 0.346 & 49.145\\ 
		ONS SOC 2010 & 3 &  0.225 & 42.665\\  		
		ONS SOC 2010 & 4 & 0.146 & 21.047\\ 
		\hline
		ONS SOC 2020 &  1  & 0.468 & 52.435 \\
		ONS SOC 2020 &  2  & 0.319 & 49.119 \\ 
		ONS SOC 2020 &  3  & 0.121 & 14.819 \\ 
		ONS SOC 2020 &  4  & 0.059 & 6.269 \\ 
		\hline
		O*NET SOC 2019 & 1  & 0.414 & 47.015 \\ 
		O*NET SOC 2019 & 2  & 0.323 & 40.762 \\ 
		O*NET SOC 2019 & 3  & 0.211 & 35.684 \\ 
		O*NET SOC 2019 & 4  & 0.194 & 27.551  \\ 
		\hline
	\end{tabular}
	\label{fig:compendiumstandaloneresult}
	\caption{Test results, derived by training models on the official 'clean' title set introduced in section \ref{datasetorig} and then performing test predictions for the 'unclean' job advertisement title dataset.    }
\end{table}



\subsection{Individual Model Performance}

Table \ref{tab:individual} summarizes the test performance (F1 Macro and Accuracy) of various individual models evaluated on the final level classification task for the ONS SOC 2020 dataset. These models explored the effectiveness of different feature sets and architectures.

As expected, the BERT Classifier model consistently achieved the best performance across both feature sets (job titles and descriptions).  This is likely due to the pre-trained nature of the BERT model, which provides a strong foundation for text classification tasks.  During training, the classification head within the BERT architecture undergoes fine-tuning specifically for our occupational classification problem, further enhancing its performance.

Here's a breakdown of some key observations from the table:

\begin{itemize}

   \item Baseline Model: This basic model utilizing only job titles as features serves as a reference point. As anticipated, its performance is relatively low (F1 Macro: 0.069, Accuracy: 23.212\%).
 \item  OvR Binary vs. Simple Models: The One-vs-Rest (OvR) Binary model, which decomposes the multi-class problem into a series of binary classifications, shows some improvement over the baseline. However, the Simple model, which directly tackles the multi-class problem, outperforms both the OvR Binary and Baseline models across both features (titles and descriptions). This suggests that the Simple model captures the inherent relationships between occupational categories more effectively.
 \item  Multi-head Attention vs. BERT Classifier: The Multi-head Attention model, while exhibiting some promise, falls short of the BERT Classifier in terms of performance. This highlights the advantage of pre-trained language models like BERT, which incorporate contextual information beyond simple word relationships for superior text classification capabilities.

\end{itemize}

\begin{table}[H]
	\centering
	\begin{tabular}{ |p{4cm}|p{3cm}||p{3cm}|p{4cm}|  }
		\hline
		\multicolumn{4}{|c|}{Individual Model Performances} \\
		\hline
		Model & Feature & Test F1 (Macro) & Test Accuracy (\%) \\
		\hline
		Baseline  & Title &  0.069 & 23.212\\
		OvR Binary  & Title & 0.168 & 50.155\\
		Simple  & Title & \textbf{0.356} & 55.026\\
		Multi-head Attention & Title  &  0.312 & 50.725\\
		BERT Classifier & Title& 0.299 & \textbf{58.135} \\
		\hline
		Baseline  & Description & 0.069 & 25.751\\
		OvR Binary  & Description & 0.082 &  34.611\\
		Simple  & Description & 0.203 & 40.259\\
		Multi-head Attention & Description  & 0.173 &  36.995 \\
		BERT Classifier & Description & \textbf{0.228} & \textbf{49.896} \\
		\hline
	\end{tabular}
	\caption{\label{tab:individual}Individual model test performances ONS SOC 2020 dataset - final level classification. Best results emphasized.}
\end{table} 

Overall, the results in Table \ref{tab:individual} emphasize the effectiveness of the BERT Classifier model and the value of incorporating rich text data (job descriptions) alongside job titles for improved occupational classification accuracy.

\subsection{Preprocessing Performance}

 The preprocessing steps employed are detailed in Section \ref{preprocessing}. As the table reveals, the impact of preprocessing on model performance is nuanced.  For job titles, preprocessed data resulted in a slight improvement in both F1 Macro (0.299 to 0.297) and Accuracy (58.135\% to 58.290\%).  This suggests that the preprocessing techniques, such as tokenization and stemming, might have a minimal positive effect on the model's ability to extract relevant information from job titles.

However, a contrasting trend emerges when examining job descriptions.  Preprocessing job descriptions led to a decrease in both F1 Macro (0.228 to 0.194) and Accuracy (49.896\% to 48.290\%).  This unexpected finding contradicts the initial expectation that preprocessing would enhance model performance.  A potential explanation for this observation lies in the nature of job descriptions.  These descriptions often contain informal language, grammatical variations, and industry-specific jargon.  While preprocessing techniques might aim to standardize the text, they might inadvertently remove some of the nuances and idiosyncrasies that carry valuable information for occupational classification.

\begin{table}[H]
	\centering
	\begin{tabular}{ |p{5cm}|p{3cm}|p{4cm}|  }
		\hline
		\multicolumn{3}{|c|}{Pre-processing Performance} \\
		\hline
		Feature & Test F1 (Macro) & Test Accuracy (\%) \\
		\hline
		Title &  \textbf{0.299} & 58.135 \\
		Title (Preprocessed) & 0.297 &  \textbf{58.290} \\
		\hline
		Description & \textbf{0.228}  & \textbf{48.896} \\
		Description (Preprocessed) & 0.194 & 48.290\\
		\hline
	\end{tabular}
	\caption{Performance of BERT classifier against raw and preprocessed data. Best results emphasized. }
\end{table} 

In light of these findings, future modeling efforts will focus on utilizing the raw data, particularly for job descriptions, to preserve the richness of the textual content and potentially improve classification accuracy.  This observation underscores the importance of careful evaluation when applying preprocessing steps, as they might not always yield the anticipated positive effects, especially when dealing with complex textual data like job descriptions.

\subsection{Hierarchical Modelling}

In section \ref{HierarchyLCPN}, we explored the possibility of using LCPN,a routing system for predictions, utilizing a series of more specific models for each "first level tree" node within the occupational taxonomy.  This system aimed to leverage an initial model to predict a first-level classification.  If the prediction confidence exceeded a predefined threshold, the sample would be routed to a specialized model trained on data points falling under that specific sub-tree.  Alternatively, if the initial prediction confidence fell below the threshold, the sample would be directed to a "flat" model capable of predicting any label (leaf node) in the entire taxonomy.

\begin{figure}[H]
	\includegraphics[width=12cm]{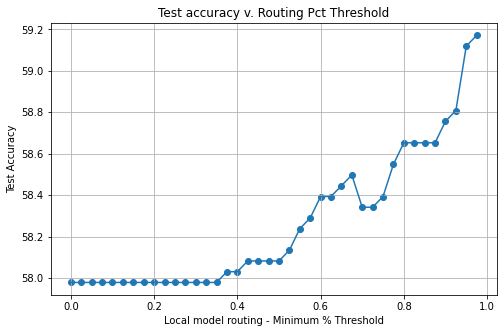}
	\caption{The result of setting different minimum pct. thresholds as a precursor to routing to a localized model.}
	\label{fig:routingthresholdtest.png}
\end{figure}

Figure \ref{fig:routingthresholdtest.png} depicts the results of setting different minimum confidence thresholds for routing to localized models.  As the figure illustrates, the model performs best with a confidence threshold of 100\%, essentially routing all samples back to the local classifier.  This outcome aligns with our observations in Section \ref{finalResult}, where the model's accuracy exhibits minimal decline during the first level of the taxonomy, with the most significant drop occurring at the final, granular level.

While we don't completely dismiss this routing approach for future endeavors, its effectiveness appears to be constrained by the current dataset size.  With a larger corpus of training data, future investigations could explore one-vs-many models built specifically for each final classification label, along with a fallback "flat" model for cases where the prediction fails to meet any confidence threshold.  Given the current limitations, however, we will not pursue the LCPN routing system for tree-based classification in this project.

Section \ref{model} discussed the potential benefits of incorporating a post-processing step that leverages knowledge about the occupational taxonomy to enhance predictive results.  Table \ref{fig:hyperfig} summarizes the outcomes of these experiments.
\begin{table}[H]
	\centering
	\begin{tabular}{ |p{7cm}||p{3cm}|  }
		\hline
		\multicolumn{2}{|c|}{Output Post-Processing Results} \\
		\hline
		Approach& Test Accuracy \% \\
		\hline
		Baseline (None)   & 58.183\\
		Total Avg.   &  58.187\\
		Weighted Avg.   &  58.394\\
		\textbf{Joint Probability}   &  \textbf{58.601}\\
		Total Avg w. Logical Pruning & 52.431\\
		Weighted Avg w. Logical Pruning & 52.435\\
		Joint Prob Avg w. Logical Pruning & 52.539\\
		\hline
	\end{tabular}
	\caption{Result of post-processing of output results. Best results emphasized. }
	\label{fig:hyperfig}
\end{table}

Various post-processing strategies were evaluated, including simple averaging, weighted averaging, and joint probability approaches.  The "Joint Probability" technique yielded the best overall accuracy (58.601\%), both with and without logical pruning (a rule-based approach to adjusting probabilities based on the taxonomy structure).  Interestingly, logical pruning did not provide any additional improvement in this case.

It's important to note that these post-processing techniques were evaluated independently, focusing on optimizing each approach in isolation.  While the "Joint Probability" method emerged as the most effective in this context, further research might be necessary to determine the optimal combination of post-processing strategies in a broader framework.

\subsection{Feature and Augmentation performance}

Our attempts at data augmentation resulted in mixed effects.  As shown in Table \ref{fig:augfig}, augmenting job titles, with skills, actually had a negative impact on test accuracy (58.134\% baseline vs. 56.943\% with augmentation).  This is likely attributable to the relatively short sequence length of job titles and the lack of domain-specific augmentation techniques.  The newly generated samples might have introduced too much similarity to existing data, tightening class boundaries and leading to class overfitting.

\begin{table}[H]
	\centering
	\begin{tabular}{ |p{3cm}||p{4cm}||p{3.5cm}|  }
		\hline
		\multicolumn{3}{|c|}{Data Augmentation - Performance Comparisons} \\
		\hline
		 Feature & Base Test Acc (\%) & Aug. Test Acc (\%)  \\
		\hline
		Job Title  &   \textbf{58.134} & 56.943 \\
		\hline
		Job Description   & 49.896 & \textbf{49.948} \\
		\hline
	\end{tabular}
	\caption{Comparison of final test performance between models trained on original dataset and augmented dataset.}
	\label{fig:augfig}
\end{table}
In contrast, augmenting job descriptions yielded a marginal improvement in test accuracy (49.896\% baseline vs. 49.948\% with augmentation).  However, the gain is not substantial enough to justify the additional complexity of data generation and integration for cross-validation.  This observation might be explained by the larger size of the text sequence in job descriptions.  The model likely relies less on individual tokens for these longer sequences, allowing for some added noise from augmentation to enhance generalization capabilities rather than causing overfitting.

\subsection{BERT Variations - Performance comparison.}

In our exploration of various pre-trained language models (PLMs) for occupational classification, Table \ref{tab:BERTvar} summarizes the trade-off between model complexity, training time, and achieved accuracy. Interestingly, larger PLMs such as RoBERTa and DeBERTa, while requiring significantly longer training times (1291 and 1304 seconds respectively) compared to DistilBERT (528 seconds), did not yield substantial improvements in test accuracy. Notably, DeBERTa achieved the lowest accuracy (54.508\%) amongst the evaluated models. These findings suggest that, in the context of our dataset with a limited number of training samples and a relatively large number of final occupational classes, the increased complexity of larger PLMs might not be fully utilized by the model. We speculate that the model may be encountering difficulties in learning effective representations for the diverse occupational categories due to insufficient training data. This observation highlights the importance of considering both model performance and computational efficiency when selecting PLMs for tasks with limited training data, particularly when dealing with a large number of classification categories.
\begin{table}[H]
	\centering
	\begin{tabular}{ |p{2cm}||p{3cm}||p{4cm}|  }
		\hline
		\multicolumn{3}{|c|}{BERT Embedding - Performance Comparisons} \\
		\hline
		 Model & Train Time (secs) & Test Accuracy (\%) \\
		\hline
		DistilBERT  &    \textbf{528} & 57.098 \\
		BERT   & 1155 & \textbf{57.150} \\
		RoBERTa  & 1291 & 56.114\\
		DeBERTa  & 1304 & 54.508\\
		\hline
	\end{tabular}
	\caption{Comparison of BERT variations computational and statistical performance. \textit{(It should be noted that training time is highly dependent on batch sizes and available hardware resources. More important are the results relative to each other.)}}
	\label{tab:BERTvar}
\end{table}

\subsection{Final Model Performance}\label{finalResult}

Tables \ref{tab:finalresult} summarize the final performance of the model on the held-out test set, consisting of 850 samples. Here, we delve into the presented metrics and discuss the impact of model pruning (specifically, leaf pruning at Tier 4). The models are analysed using the metrics: 

\begin{itemize}
    \item F1 Score (Macro): This metric is employed to account for imbalanced class distributions within the dataset (as visualized in Figure \ref{fig:soccodejobfig}). It provides a harmonic mean of precision and recall, offering a balanced view of model performance across all classes, even those with fewer samples.
 \item T-1, T-5, T-10 Accuracy: These metrics represent the percentage of samples where the correct classification falls within the top 1, top 5, and top 10 predictions, respectively, when the model's outputs are ordered by their predicted confidence scores.
\item Level: This refers to the hierarchical level within the occupational taxonomy being predicted. Level 1 represents the broadest classification, while Level 4 corresponds to the most granular (leaf) nodes. It's important to note that for T-10 accuracy at Level 1, results are not shown for the ONS datasets because these datasets only have 10 top-level groups. In such cases, the true classification will always be included within the top 10 predictions by definition.
\end{itemize}

\begin{table}[H]
\begin{center}
	\begin{tabular}{ |p{3cm}|p{1cm}|p{1.2cm}|p{1cm}|p{2cm}|p{2cm}|p{2.2cm}|  }
	\hline
	\multicolumn{7}{|c|}{Final Model Test Performance - Without Leaf (Tier 4) Pruning} \\
	\hline
	Classification & Level & Labels & F1 & T-1 acc (\%) & T-5 acc (\%) & T-10 acc (\%) \\
	\hline
	ONS SOC 2010 &  1 & 10& 0.713 & 78.434 & 97.149 & - \\ 
	ONS SOC 2010 &  2 & 28& 0.558 & 74.235 & 92.846 & 96.216 \\ 
	ONS SOC 2010 &  3 & 100& 0.434 & 71.125 & 87.195 & 90.876 \\ 
	\textbf{ONS SOC 2010} & \textbf{4} &  \textbf{369} &\textbf{0.325} & \textbf{61.223} & \textbf{81.908} & \textbf{86.159} \\ 
	\hline
	ONS SOC 2020 &  1 & 10& 0.701 & 76.891 & 96.943 & - \\ 
	ONS SOC 2020 &  2 & 31& 0.535 & 73.057 & 91.554 & 96.269 \\ 
	ONS SOC 2020 &  3 & 122& 0.405 & 70.259 & 86.580 & 90.777 \\ 
	\textbf{ONS SOC 2020} &  \textbf{4} & \textbf{412} &\textbf{0.312} & \textbf{59.170} & \textbf{80.207} & \textbf{84.456} \\ 
	\hline
	O*NET 2019 &  1 & 24& 0.571 & 69.394 & 92.619 & 97.743 \\ 
	O*NET 2019  &  2 & 100& 0.427 & 59.426 & 85.331 & 91.255 \\ 
	O*NET 2019  &  3 & 462 & 0.304 & 54.067 & 80.536 & 85.002 \\ 
	\textbf{O*NET 2019} & \textbf{4} & \textbf{867} & \textbf{0.230} & \textbf{44.993} & \textbf{74.659} & \textbf{81.335} \\ 
	\hline
\end{tabular}
\caption{Final Model Results without any pruning performed at tier 4 }
\label{tab:finalresult}
\end{center}
\end{table}

To achieve the reported performance, a total of 24 separate models were trained. Each model focused on a specific classification level (Level 1 to 4) within a particular occupational dataset (ONS SOC 2010, ONS SOC 2020, or O*NET 2019). Additionally, each model was trained on both job titles and job descriptions to leverage the combined information for improved classification. These individual models then participated in an ensemble prediction scheme, where their outputs were combined using a technique called "combined probability pruning."

The overall lower F1 scores can be attributed to the utilization of macro-averaging for the F1 metric. This approach is particularly well-suited for handling imbalanced datasets, where some classes have significantly fewer samples compared to others. The data exhibits such an imbalance, with certain occupational categories having limited representation. The F1 score reflects the model's performance on these under-represented classes, leading to a lower overall score.

\begin{table}[H]
\begin{center}

	\begin{tabular}{ |p{3cm}|p{1cm}|p{2cm}|p{2cm}|p{2.2cm}|  }
		\hline
		\multicolumn{5}{|c|}{Final Model Test Performance - With Leaf (Tier 4) Pruning} \\
		\hline
		Classification  & F1 & T-1 acc (\%) & T-5 acc (\%) & T-10 acc (\%) \\
		\hline
		ONS SOC 2010 & 0.303 & 58.165 & 74.028 & 79.368 \\ 
		ONS SOC 2020 &  0.312 & 60.259 & 75.648 & 80.104 \\ 
		O*NET SOC 2019 & 0.270 & 46.262 & 73.061 & 78.561 \\ 
		\hline
	\end{tabular}
 \caption{Model results with pruning}
	\end{center}
\end{table}

The effectiveness of joint probability pruning appears to be dataset-dependent. While it leads to an increase in top-1 accuracy for the ONS SOC 2020 and O*NET 2019 datasets, this gain comes at the expense of a 1-2\% decrease in top-5 and top-10 accuracy for these datasets. Furthermore, for the ONS SOC 2010 dataset, joint probability pruning results in a 2-3\% decline in accuracy across all T-levels. This observation suggests that joint probability pruning might have the unintended consequence of widening the classification boundaries. While it boosts confidence in the most likely classification, it might also push other potential classifications further away, leading to a decrease in accuracy for less-confident predictions, particularly for the top-5 and top-10 metrics.

\subsection{Error Analysis}
Upon examining the results presented in Table \ref{tab:finalresult}, a discernible relationship emerges between the granularity of classification levels and the deterioration in test accuracy. Specifically, there is an almost universal 10\% decline in predictive accuracy observed when transitioning from tier 3 to tier 4 (the final classification level). This phenomenon can be attributed to the diminution in sample size attendant upon each successive, more nuanced subdivision of the label space. A correlation between sample size and accuracy is evident and aligns with theoretical expectations, given that larger sample sizes tend to provide more robust statistical bases for predictive modeling.

\begin{figure}[H]
	\centering
	\includegraphics[width=10cm]{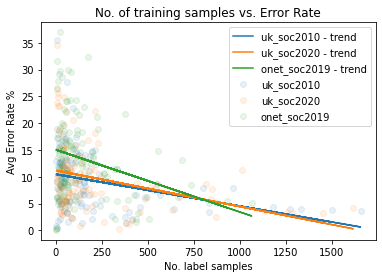}
	\caption{Error rate with labelled data }
	\label{fig:samplecorr}
\end{figure}

In the context of the O*NET dataset performance, the graphical representation in Figure \ref{fig:samplecorr} elucidates the inverse relationship between error rates and the volume of labeled data. 

\begin{figure}[H]
	\centering
	\includegraphics[width=6cm]{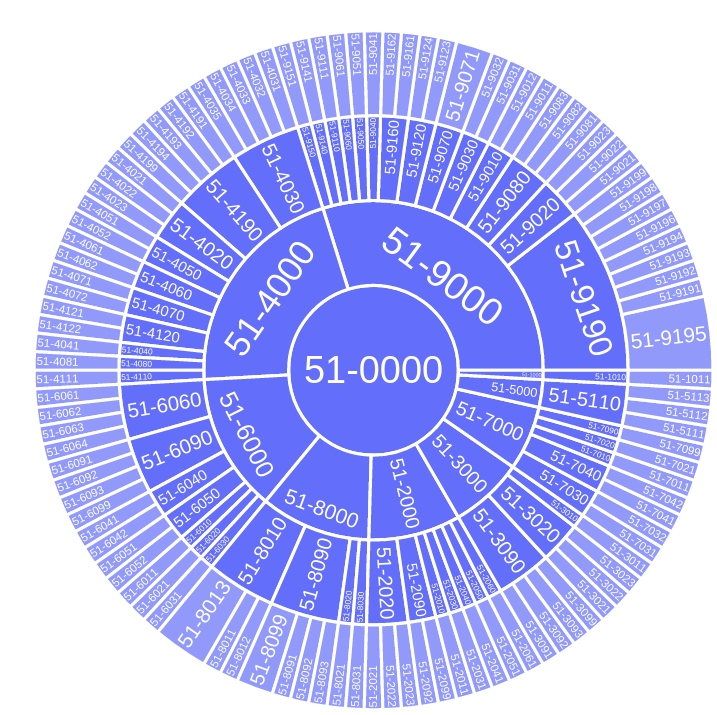}
	\caption{One of the 23 major groups of the O*NET 2019 standard, demonstrating the severe proliferation of classifications.  }
	\label{fig:O*NETaaa}
\end{figure}

Furthermore, Figure \ref{fig:O*NETaaa} showcases the proliferation of classifications within one of the 23 major groups of the O*NET 2019 standard, highlighting the severe fragmentation of categories. This expansive diversification of classifications under the O*NET standard, which possesses nearly quadruple the number of tree nodes relative to the ONS SOC taxonomy, introduces considerable complexity into the predictive task. The subdivision of sample sets for final level classification becomes increasingly pronounced, leading to a reduction in the statistical power of the models.

This granular subdivision poses a substantial challenge, as elucidated through the examination of confusion matrices in section \ref{confmatrices}, which reveal no significant large-scale misclassification of test samples into any particular group. The analysis primarily concentrates on the major group classifications, predicated on the assumption that these groups possess adequate sample sizes to ensure accurate fitting. The blocking issue noted herein implies that accuracy at this broader level of classification taxonomy has profound implications for the precision of more detailed classifications.

The empirical evidence suggests that while increasing the specificity of classification can enhance the granularity of labor market analysis, it concomitantly imposes limitations on predictive accuracy due to reduced sample sizes and increased categorization complexity. This finding underscores the inherent trade-off between classification detail and predictive efficacy, which is a critical consideration for economists and policymakers utilizing such data for labor market analysis.

\section{Conclusion}

The endeavor to assign Standard Occupational Classification (SOC) codes to job advertisements is a pivotal task for enhancing labor market analysis and facilitating effective job matching. This review has scrutinized existing methodologies for automating this assignment process, drawing from a selection of seminal works in the field.

Textual analysis emerges as a common thread among the discussed methods, with studies employing various strategies from rule-based approaches to advanced machine learning models to interpret and classify job postings accurately. The Continuous Bag-of-Words model explored by \cite{atalay2020evolution} and the rule-based method employed by \cite{turrell2019transforming} underscore the diversity of techniques available for tackling this classification challenge. Each approach, while unique in its methodology, underscores the inherent trade-offs between classification granularity and accuracy.

Supervised machine learning models, as demonstrated by \cite{gweon2017three} and \cite{boselli2018wolmis}, further attest to the potential of computational methods in enhancing the precision and scalability of SOC code assignments. These studies reveal the efficacy of incorporating rich textual features and leveraging comprehensive training datasets for improved classification outcomes.

This paper contributes to the discourse by integrating the advanced capabilities of BERT transformers for the classification of raw job advertisements into UK SOC and O*NET SOC codes. The ensemble prediction scheme, underpinned by joint probability pruning, marks a novel approach in this research domain. However, the effectiveness of this methodology appears contingent upon the dataset characteristics, with its impact varying across different classification levels and SOC datasets. The findings suggest that while there is room for optimization, particularly in addressing dataset-specific challenges and enhancing the robustness of predictions for under-represented classes, the employed techniques hold substantial promise for the future of automated job classification.

In comparison to existing methodologies, our model demonstrates a competitive edge in accuracy across a comprehensive sample of job advertisements. While the performance of \cite{turrell2019transforming}(e.g., 91\% accuracy for vacancy postings on a subset of classifiable ads) and \cite{boselli2018wolmis}(88\% with a training set 5 times bigger) is noteworthy, our approach achieves commendable accuracy even within the constraints of smaller training sets and broader sample inclusivity. The \cite{turrell2019transforming} model was implemented and evaluated using the same training dataset employed in this study.  This implementation achieved a predictive accuracy of 33\% at the UK SOC 2010, only classifying 67\% of job advertisements.  However, when considering only the subset of job ads for which the model produced a prediction, the accuracy increased to 83\%, consistent with their reported high accuracy on selected samples.  In contrast, the present model achieved a prediction rate of 71\% across the entire dataset without any pre-selection.  Further analysis revealed that the decline in \cite{turrell2019transforming} accuracy is correlated with job title length.   

Our method achieves high accuracy, exceeding 80\% for top-5 predictions even at the most granular (Level 4) classification. While the overall F1 score is lower due to the chosen metric's sensitivity to imbalanced datasets, the model demonstrates strong performance across various occupational categories. In conclusion, the exploration of automated SOC code assignment to job advertisements reveals a landscape rich with potential yet fraught with challenges. The studies reviewed herein collectively pave the way for future innovations in this field, highlighting the importance of methodological diversity, dataset considerations, and the ongoing need for refinements to achieve both high granularity and accuracy in classification tasks. As we look forward, it is clear that the expansion of training datasets and the continued evolution of machine learning models will be crucial in surmounting current limitations and unlocking the full potential of automated job classification systems.

\bibliographystyle{plainnat}
\bibliography{ref}
\newpage
\appendix

\section{Language Models}

\subsection{BERT}

BERT stands for Bidirectional Encoder Representations from Transformers  \citep{RogersAnna2021APiB}. It's a powerful machine learning model trained on massive amounts of text data. BERT can "read" text and understand the relationships between words,  allowing it to interpret the meaning within a job advertisement even if the language is informal or contains typos. Both BERT and GPT-3 are powerful language models, but they have different strengths and weaknesses. BERT excels at understanding existing text data, while GPT-3 is better at generating creative text formats. Our research focuses on classification tasks, making BERT a more suitable choice \citep{gaikwad2022extensive}.

\begin{figure}[H]
	\includegraphics[width=15cm]{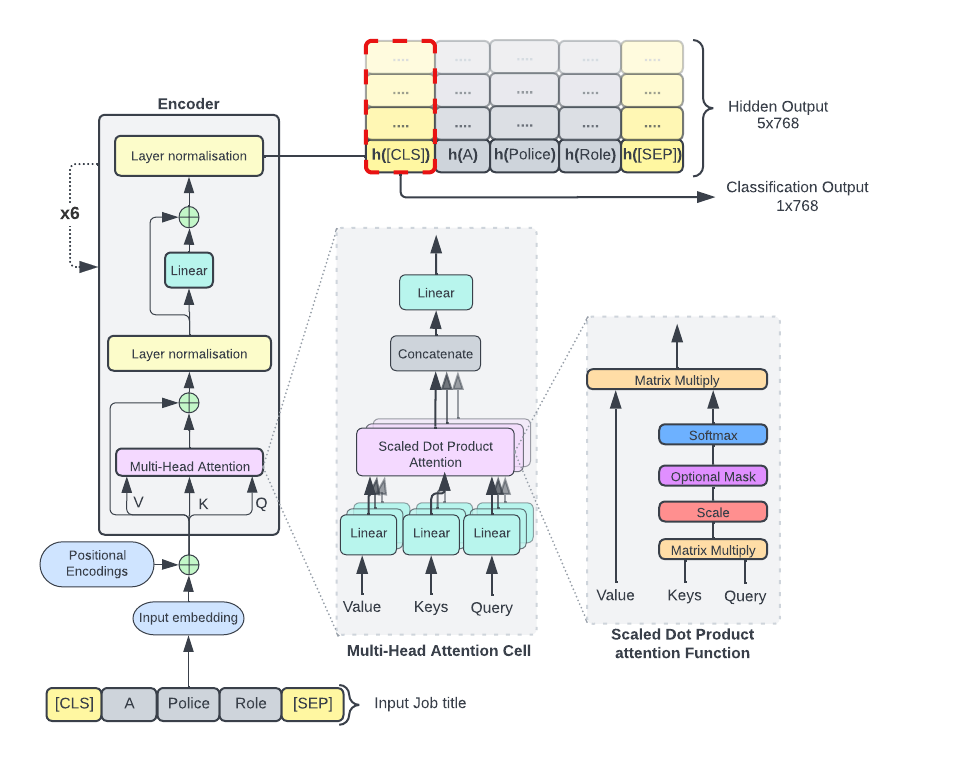}
	\caption{Example of Bert encoder, showing the breakdown of the attention functionality -  adapted from examples provided  \citep{VaswaniAshish2017AIAY} note the use of the 'classification' token [CLS]- used instead of an encoder for classification based tasks. }
	\label{fig:bertle}
\end{figure}

BERT's architecture (Figure \ref{fig:bertle})revolves around two primary components: the encoder and (optionally) the decoder. The encoder plays a crucial role in capturing contextual relationships between words and sub-words within a given text. This is achieved through a mechanism called "self-attention," \citep{VaswaniAshish2017AIAY}. Self-attention allows the model to analyze interactions between all tokens (words) and sub-tokens within a sentence simultaneously \citep{DevlinJacob2018BPoD}. This permits the model to assign weights (importance scores) to each token based on its relevance to the overall meaning of the sentence. These weights are calculated for every possible pairing of tokens, resulting in a final output vector containing weights for all n input tokens. This approach empowers BERT to effectively capture context by considering the entire sentence at once, overcoming limitations associated with sequential processing methods. Additionally, it mitigates the issue of "catastrophic forgetting," a phenomenon where historical information is progressively lost in longer sequences.

To derive contextually-aware embeddings for each token during the encoding phase, BERT employs two complementary techniques. The first, Masked Language Modeling (MLM), involves replacing a certain number of tokens with a special "[MASK]" token. The model then endeavors to predict the masked tokens based on the surrounding context provided by the unmasked tokens. This process essentially trains BERT to create embeddings that are sensitive to the surrounding context.  In contrast to traditional NLP models that process tokens sequentially, BERT analyzes the entire sequence simultaneously, enabling it to embed context with respect to the whole sentence. This not only enhances performance but also avoids issues with longer sequences.

In addition to MLM, BERT utilizes Next Sentence Prediction (NSP) as a supplementary training approach. During NSP, pairs of sentences are presented as input, with the first sentence followed by either the actual subsequent sentence or a random sentence (50\% chance). The model then attempts to predict if the second sentence logically follows the first. Both MLM and NSP contribute to minimizing a combined loss function during training, ultimately enabling BERT to refine its ability to identify contextual relationships within text.

The ability of BERT to be fine-tuned in an unsupervised manner using vast amounts of unlabeled data presents a significant advantage. This allows models like RoBERTa \citep{roberta}, trained on a massive dataset of books, articles, web pages, and other sources, to leverage the power of unlabeled information. This unsupervised pre-training empowers models to acquire a robust understanding of language before being fine-tuned for specific tasks like occupational classification in job advertisements.

The decoder component, when used, receives the encoded information for further processing, typically leading to the generation of an output sequence. However, for occupational classification tasks, we can bypass the decoder phase and instead utilize the hidden representation of the "[CLS]" classification token, which effectively encapsulates the meaning of the entire encoded sentence. This hidden representation is then fed through a linear layer and a softmax layer to assign probabilities across the output vocabulary (occupation categories in our case).

While pre-trained models like "BertForSequenceClassification" already incorporate a "BertPooler" layer to extract the hidden representation of the "[CLS]" token, our approach involves manually extracting this information during data preprocessing. This ensures that the prepared data can be readily used for a wider range of model experiments without modifications. By leveraging BERT's unique architecture and its ability to capture context through self-attention and pre-training on massive datasets, we can achieve superior performance in classifying occupations based on job advertisement text data.

\subsection{Variations of BERT}

While the base BERT model offers undeniable power, its training requirements can be substantial, with the large variant boasting 340 million parameters.  This necessitates exploring computationally efficient alternatives for rapid experimentation. Our initial investigation focuses on DistilBERT, a streamlined version of BERT that leverages knowledge distillation.

Knowledge distillation \citep{HintonGeoffrey2015DtKi}, operates under the principle that a model's final output (including BERT's softmax distribution) can be utilized to train a new, "student" model. Softmax distributions represent the probabilities assigned to various categories by the model. The student model is trained on a transfer dataset, aiming to minimize the cross-entropy between its own output distribution and that of the pre-trained "teacher" model (BERT in this case). Mathematically, this translates to minimizing the expression:

$$ \sum_{i} \hat{y}_i(x|t) \log(y_i(x|t)) $$
where: \\ 
\begin{center}

$t$  represents a "temperature" value influencing the output entropy. \\
$\hat{y}_i(x|t)$  signifies the original model's (teacher's) distribution.\\
$y_i(x|t)$  denotes the distilled model's (student's) distribution.\\
\end{center}
Combined, these elements represent the probability distribution produced by the softmax function, with higher temperature values "softening" the distribution (increasing entropy). Distillation essentially allows a smaller model to learn from a larger, pre-trained model, achieving a balance between performance and computational efficiency.

In practice, DistilBERT leverages knowledge distillation to create a more computationally efficient version of BERT, capturing approximately 95\% of its performance while requiring 60\% fewer computations \citep{SanhVictor2019Dadv}. This makes DistilBERT an ideal starting point for our exploration. Following a thorough investigation of DistilBERT's performance in the context of occupational classification, particularly with regard to hierarchical classifications and model pruning techniques, we will progress to exploring larger embedding models for our final model(s). Here's an overview of the potential candidates:

\begin{itemize}

 \item  BERT Base: The previously described BERT model.

 \item  RoBERTa: A variant of BERT developed by Meta Platforms, Inc. (formerly Facebook). RoBERTa focuses on retraining BERT with a significantly larger (roughly tenfold) dataset and meticulously optimized hyperparameters. This approach aims to generate more contextually-aware embeddings.

 \item  DeBERTa:  DeBERTa is a Microsoft-developed variation of BERT that seeks to improve upon both BERT and RoBERTa \citep{HePengcheng2020DDBw}. It achieves this through two key innovations: Disentangled Attention Mechanism: Instead of combining content and position vectors, DeBERTa keeps them separate. This allows for self-attention to be computed based on both the semantic content of words and their relative position within a sentence. Incorporating Disentangled Representations: The model then utilizes these separate content and position representations during the decoding process. This facilitates a deeper understanding of the various syntactic roles words play within sentences.
\end{itemize}

By systematically evaluating these models, we can determine the most effective architecture for our specific task of occupational classification from job advertisement text data.

\section{Large Document Analysis}
\subsection{Flat classification}\label{Flatclass}

Our initial exploration focuses on "flat" classification architectures, aiming to directly predict the most granular occupational category for a given job advertisement.  This approach offers the advantage of structural simplicity, requiring minimal customization of pre-existing components. However, it  disregards the inherent hierarchical structure present within occupational classification taxonomies.  We will compare the performance of various multi-class models within this flat classification framework:

\begin{itemize}

 \item  Baseline: This baseline model employs a single classification layer directly atop the output embeddings generated by the BERT encoder.

 \item  Simple Model: This model incorporates a classification head on the BERT encoder output. This head includes a linear layer, layer normalization, dropout regularization, a GELU (Gaussian Error Linear Unit) activation function, and finally a single output layer for classification \citep{hendrycks2016gaussian}.)

 \item  Multihead Attention Model: This model utilizes four layers, each consisting of a multi-head attention mechanism with 12 attention heads and a linear layer, similar to the "Simple" model. These layers are followed by a single final classification layer.

 \item  One-vs-Rest Binary Model (OVM Binary): This model shares similarities with the "Simple" model but differs in its output layer.  The OVM Binary model predicts a single binary value wrapped in a sigmoid activation function.  A separate model is built for each possible occupational category, facilitating one-versus-rest binary predictions.

 \item  BERT Classifier: This model leverages the pre-trained classification head provided with the BERT model. This head typically comprises several multi-head attention layers, each containing multiple linear layers, GELU activations, and layer normalization steps.

\end{itemize}

By evaluating these diverse flat classification models, we aim to establish a baseline performance benchmark for occupational classification using BERT on job advertisement text data. This initial assessment will inform subsequent investigations into more elaborate hierarchical classification approaches that exploit the inherent structure within occupational taxonomies.
 
\subsection{Hierarchical Classification}
\label{HierarchyLCPN}

This section delves into the strengths and limitations of various approaches for hierarchical classification of occupations using BERT embeddings. We differentiate these hierarchical approaches from the previously discussed flat classification methods, which do not leverage the inherent structure within occupational taxonomies. The sheer number of potential occupational categories and the inherent complexity involved in precisely identifying granular job classifications motivate us to explore hierarchical classification approaches.  Hierarchical classification frameworks leverage the inherent structure within occupational taxonomies, potentially utilizing more accurate higher-level predictions to inform and refine lower-level classifications.  Advancements in computational efficiency  propose a method that utilizes multiple classifiers, each trained on specific subsets of the taxonomic tree \citep{SillaCarlosN2010Asoh}. This approach enables individual classifiers to specialize in localized regions within the overall classification hierarchy.

Here, we delve into five distinct hierarchical classification variants applicable to our occupational classification task using BERT embeddings:

\begin{itemize}

    \item  Flat Classifier (FC): This classifier directly predicts the final leaf nodes in the occupational taxonomy – the most specific job occupation codes in our case.  This approach, discussed in the previous section , serves as a baseline for comparison with hierarchical methods.

    \item  Global Classifier (GC): In contrast to the FC approach, the GC attempts to predict any possible node within the taxonomy, irrespective of its level.  This offers greater flexibility but may come at the cost of reduced accuracy, particularly for deeper levels within the taxonomy.

    \item  Local Classifier per Node (LCN): This approach employs a binary classifier for each node in the taxonomy, aiming to determine if a job advertisement belongs to that specific node. While a powerful approach, LCN models require a significant number of training samples to build robust classifiers for each node, especially considering the large number of leaf nodes in occupational taxonomies.  Given the limitations of our dataset size, LCN is not a viable option for this project.

    \item  Local Classifier per Parent Node (LCPN): This method utilizes a multi-class classifier for each parent node in the hierarchy. The objective is to predict the child node, under a specific parent node, to which a given job advertisement belongs.  LCPN offers a balance between accuracy and computational efficiency compared to the other hierarchical approaches.  Furthermore, it allows us to selectively train classifiers only for branches with sufficient training data, while utilizing a more general classifier for branches with limited data.  We can leverage the model's confidence score in its predictions to determine whether a sample should be routed to a local parent classifier (LCPN) or the flat classifier for final classification.

    \item  Local Classifier per Level (LCL): This variant employs a multi-class classifier for each level within the taxonomy.  The LCL model attempts to classify a job advertisement into the appropriate category at each level of the occupational hierarchy.

\end{itemize}

Both flat classification and LCL approaches avoid the "blocking problem"  commonly encountered in hierarchical classification tasks \citep{SunChi2019HtFB}. This problem arises when an error at a higher level in the classification hierarchy propagates down, leading to cascading misclassifications at lower levels. However, this benefit comes at a cost. Flat and LCL models  don't utilize the hierarchical structure of the occupational taxonomy, potentially sacrificing accuracy for simplicity.

Furthermore, LCL approaches can lead to inconsistencies within classifications. For instance, an LCL model might classify a job advertisement under a specific major occupational group (e.g., ONS SOC code 5000) but then assign it to a sub-major group (e.g., ONS SOC code 4800) that falls outside of that major group (correct code should be within the range of 5000-5999). Additionally, LCL models can be computationally expensive. As LCL essentially performs one-vs-rest classification for every leaf category, it requires invoking each classification model for final label assignment.

\cite{KiritchenkoSvetlana2006LaEi} propose treating hierarchical classification as a multi-label classification problem.  This involves expanding the label set for each sample to include all ancestor nodes along the path to the most specific classification.  Post-processing techniques can then be applied to remove inconsistencies within these expanded labels.  Additionally, they introduce hierarchical precision (hP), recall (hR), and F-score (hF) metrics that account for the hierarchical structure and penalize misclassifications to ancestor nodes less severely than misclassifications to unrelated categories.

While these evaluation metrics and the concept of treating hierarchical classification as a multi-label problem hold merit,  we prioritize final-level classification accuracy for this project.  Future research can explore the utility of these techniques in greater depth.

The hierarchical F1 score (hF) metric is not suitable for our purposes due to its non-differentiable nature, hindering its use within standard gradient-based optimization algorithms commonly employed for training deep learning models.  \cite{WuCinna2019Ahla} highlight that current back-propagation approaches struggle to effectively "tie together" ancestral errors during training in hierarchical classification tasks.  This limits the effectiveness of leveraging hierarchical structure within the objective function itself.  While this area of research holds promise for future advancements in hierarchical classification, it falls outside the scope of this project.

\section{Model}
\label{model:appendix}

Our initial analysis used the BERT model to process job advertisement text and create meaningful numerical representations (embeddings) of the data. These embeddings were then fed into different types of neural networks to classify job postings into occupational categories. We experimented with various neural network structures, adjusting their size, activation functions, and techniques to prevent overfitting. To train the model effectively, we used a cross-entropy loss function, which helps distinguish between multiple occupational categories. Additionally, we tested a one-vs-rest (OVR) approach, where we used a binary classifier for each top-level occupation. However, this method consistently performed worse than other models.

Interestingly, none of the custom-built neural networks outperformed a pre-trained BERT model (BertForSequenceClassification) from the Transformers library. This model relies on multiple transformer layers, each containing a feed-forward component and a self-attention mechanism. The self-attention layer is particularly useful for understanding text, as it highlights important parts of a sentence while downplaying less relevant information.

To address challenges such as unstable gradients, we applied gradient clipping, which limits sudden changes in model weights during training. Additionally, to avoid overfitting and improve efficiency, we used early stopping—monitoring validation accuracy and halting training if no improvement was seen over a set number of training cycles.

For the "skill" classification task, the best performance came from a relatively simple deep neural network with three hidden layers (512, 1024, and 512 neurons). It used GELU (Gaussian Error Linear Unit) activation functions, layer normalization, and a dropout rate of 0.3 to reduce overfitting. This finding suggests that simpler models can sometimes be sufficient for certain classification tasks.

When refining our approach, we explored a hierarchical classification method known as the Local Classifier per Parent Node (LCPN) \citep{KiritchenkoSvetlana2006LaEi}. Here, a top-level model first assigns a broad occupational category, and then specialized models handle more detailed classifications within each category. While promising, this method depends on the accuracy of the first classification step—errors at the top level can affect the entire classification tree. To address this, we set a confidence threshold: only highly confident classifications proceeded to the next level, while uncertain cases were classified by the top-level model alone. We also tested different methods for incorporating confidence scores into final predictions:

\begin{itemize}
    \item Total Average: Assigns equal weight to confidence scores from all classification levels.
    \item Weighted Average: Adjusts the weight of confidence scores based on their importance.
    \item Joint Probability: Multiplies confidence scores from each level to determine the final classification probability.
\end{itemize}

A challenge we encountered was how to handle situations where the model was confident in a final classification but uncertain about higher-level categories. To address this, we tested "logical ancestry pruning," which removes uncertain classifications to produce a more reliable final output.

\subsection{Optimizer} 

For optimization, we used AdamW, an improved version of the Adam optimizer available in PyTorch. Optimization plays a crucial role in model performance by adjusting the weights of the neural network to minimize errors and improve accuracy. AdamW refines weight decay, which helps prevent overfitting, ensuring that the model generalizes well to new data. AdamW enhances model performance by refining how weight decay is applied, leading to better generalization. Our experiments confirmed its effectiveness, particularly when combined with a learning rate adjustment technique known as cosine annealing, which gradually decreases the learning rate over time. Through extensive tuning, we found the best learning rate to be 1.26e-4 and the optimal weight decay factor to be 1.52e-6 \citep{LoshchilovIlya2017DWDR}.

\subsection{Regularization}

To prevent overfitting in our feed-forward neural networks, we initially used a dropout rate of 0.3, a common choice in deep learning. Dropout works by randomly disabling a fraction of neurons during training, which prevents the model from becoming overly dependent on specific neurons and improves its ability to generalize to new data. Additionally, we applied layer normalization, which standardizes neuron activations at each layer, ensuring that the scale of inputs remains consistent throughout the network\citep{layernorm}. This stabilization helps speed up training and prevents issues like exploding or vanishing gradients, leading to a more reliable model.

Upon transitioning to the full BERT encoder and classification head architecture, we integrated hyper-parameter tuning to optimize the dropout rates within the pretrained BERT model itself.  As previously mentioned, each transformer block within BERT consists of a feed-forward layer and an attention head, both of which incorporate a default dropout rate of 0.1.  Our hyper-parameter tuning process allowed for the exploration of a uniform dropout rate for all feed-forward layers and a separate uniform dropout rate for all attention heads within the BERT architecture.  This tuning resulted in optimal values of 0.25 and 0.35 for the feed-forward layer dropout and attention head dropout, respectively.

To mitigate overfitting and enhance training efficiency, we implemented an early stopping strategy within the training process.  This strategy monitors the validation accuracy of the model over a pre-defined patience window (e.g., 5 epochs in our case).  If the validation accuracy fails to improve within this window, the training process is terminated.  This approach prevents the model from continuing to train on data that might lead to overfitting and ultimately hinder generalizability on unseen data. By incorporating dropout rates, layer normalization, and early stopping, we aimed to achieve a balance between model complexity, training efficiency, and generalization performance in the context of occupational classification.

\subsection{Hyper-Parameter Tuning}

For optimizing the hyper-parameters of our models, we leveraged Optuna, a well-established Python library specifically designed for hyper-parameter tuning tasks.  Given the substantial computational resources required for training large language models like BERT and limitations in our available computing power, we adopted the Tree-structured Parzen Estimator (TPE) approach  \citep{Ozaki2020}.  TPE utilizes Gaussian Mixture Modeling (GMM) for efficient hyper-parameter search and aims to solve the following optimization problem:

\begin{center}

$\underset{x}{\mathrm{argmax}}  \dfrac{a(x)}{b(x)}$ ,

\end{center}

where $a(x)$ represents the hyper-parameter configuration associated with the best observed objective value, and $b(x)$ represents the remaining hyper-parameter configurations.  This formulation essentially seeks the hyper-parameter configuration that maximizes the ratio of the probability of achieving the best objective result to the probability of exploring other untested configurations.

We employed Optunawith TPE to perform 100 trials, utilizing 5-fold cross-validation to ensure robust hyper-parameter selection.  The following hyper-parameters were designated for optimization:
\begin{itemize}

    \item Learning rate ($\alpha$)
    \item $L_2$ regularization weight ($\lambda$)
    \item Number of training epochs
    \item Dropout rate for attention heads within the BERT transformer model
    \item Dropout rate for hidden layers within the BERT transformer model
\end{itemize}
The results of this hyper-parameter tuning process are detailed in the table below (Table \ref{tab:hyper}).

\begin{table}[H]
	\centering
	\begin{tabular}{ |p{4cm}||p{3cm}||p{4cm}|  }
		\hline
		\multicolumn{3}{|c|}{Hyper-parameter Ranges} \\
		\hline
		Hyperparameter& Range & Range Format\\
		\hline
		Epochs   & $ 5 \to 100$ & Multiples of 5\\
		Attention Dropout   & $ 0.1 \to 0.6$ & Multiples 0.05\\
		Hidden Layer Dropout   &  $0.1 \to 0.6$ & Multiples 0.05\\
		$\lambda$ & $\textbf{1e-9} \to \textbf{1e-2}$ & Logarithmic uniform\\
		$\alpha$ & $\textbf{1e-6} \to \textbf{1e-1}$ & Logarithmic uniform\\
		\hline
	\end{tabular}
	\caption{\label{tab:hyper1}Ranges permitted during hyper-parameter tuning}
\end{table}

\begin{table}[H]
	\centering
	\begin{tabular}{ |p{5cm}||p{3cm}|  }
		\hline
		\multicolumn{2}{|c|}{Hyper-parameter results} \\
		\hline
		Hyperparameter& Optimal value\\
		\hline
		Epochs   & 75\\
		Attention Dropout   &  0.25\\
		Hidden Layer Dropout   &  0.35\\
		$\lambda$ & 1.52e-6\\
		$\alpha$ & 1.26e-4\\
		\hline
	\end{tabular}
\caption{Final result of 100 hyper-parameter trials.}
\label{tab:hyper}
\end{table}

By employing Optuna and TPE, we aimed to achieve an optimal configuration of hyper-parameters that would lead to the best possible performance metrics for our occupational classification models, considering the constraints of our computational resources.


\subsection{Gradient Accumulation}
A significant challenge encountered during the initial training phase involved the substantial memory footprint required when training models on the "description" dataset.  The original BERT paper  \citep{DevlinJacob2018BPoD} identifies the inherent quadratic complexity ($O(n^2$) of the self-attention layer within the BERT architecture, where n represents the sequence length.  This complexity translates to a rapid increase in memory consumption as the length of the input text sequences grows.  Consequently, without any model adjustments, batch sizes during training were rarely possible beyond sixteen samples due to memory limitations.

To address this challenge and ensure a computationally feasible training process without compromising model accuracy, we implemented gradient accumulation.  This technique involves accumulating gradients across multiple mini-batches before performing a single back-propagation step.  In our case, we configured gradient accumulation to perform back-propagation every 20 mini-batches, each containing 16 samples, resulting in an effective batch size of 320 samples for training.  This approach effectively reduces the memory requirements per training step while maintaining the ability to learn from the same amount of data overall.

\subsection{Ensemble} 	
One of the key advantages of the provided dataset was the inclusion of the full occupational description, offering a richer source of information compared to prior studies (e.g., \citep{lima_bakhshi_2018}) that relied solely on job titles and extracted skill sets for occupational classification.

This expanded feature set allowed us to explore the effectiveness of various models utilizing different combinations of features: job titles, job descriptions, extracted skills, and official classifier titles.  In practice, we opted for a weighted ensemble approach, where the output scores from each individual model were combined with specific weights before applying a hard maximum operation to determine the final predicted occupational category.

The benefit of this ensemble strategy was evident in the improved classification accuracy for the final occupational level.  Compared to a model using only the job title for prediction, the ensemble approach yielded an additional 3-5\% improvement in accuracy.  While this seemingly incremental gain might appear modest, it translates to a significant enhancement in performance when considering the vast scale of occupational classification tasks.  For instance, in a scenario involving millions of job postings, a 3-5\% improvement in accuracy translates to a substantial number of correctly classified jobs, potentially leading to better targeted workforce development initiatives or more efficient labour market matching processes.

\section{Additional results}
\subsection{Final results - confusion matrices}\label{confmatrices}
\begin{figure}[H]
	\centering
	\includegraphics[width=8cm]{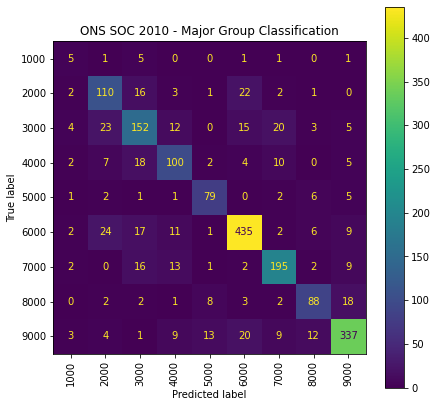}
	\caption{ONS Soc 2010 confusion matrix - first tier}
\end{figure}
\begin{figure}[H]
	\centering
	\includegraphics[width=8cm]{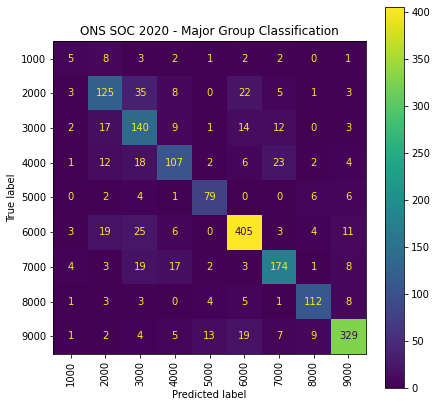}
	\caption{ONS Soc 2020 confusion matrix - first tier}
\end{figure}
\begin{figure}[H]
	\centering
	\includegraphics[width=8cm]{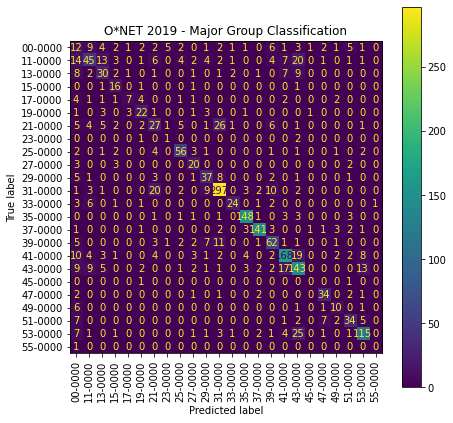}
	\caption{O*NET 2019 confusion matrix - first tier}
\end{figure}

\end{document}